\documentclass[sigconf]{acmart}

\AtBeginDocument
{
        
}

\copyrightyear{2026}
\acmYear{2026}
\setcopyright{cc}
\setcctype{by-nc-nd}
\acmConference[MM '26]{Proceedings of the 34th ACM International Conference on Multimedia}{November 10--14, 2026}{Rio de Janeiro, Brazil}
\acmBooktitle{Proceedings of the 34th ACM International Conference on Multimedia (MM '26), November 10--14, 2026, Rio de Janeiro, Brazil}
\acmDOI{10.1145/3767308.3836397}
\acmISBN{979-8-4007-2213-4/2026/11}

\usepackage{multirow}
\usepackage{tabularx}
\usepackage{amsmath}
\usepackage{multicol}
\usepackage{booktabs}
\usepackage{makecell}
\usepackage{pifont}
\usepackage{xcolor}
\usepackage{colortbl}
\usepackage{tikz}
\usepackage{graphicx}
\usepackage{enumitem}
\newcommand{\cmark}{\textcolor{green!50!black}{\ding{51}}}

\newcommand{\xmark}{\textcolor{red!60!black}{\ding{55}}}

\usepackage[most]{tcolorbox}
\tcbuselibrary{skins,breakable}
\usepackage{fvextra}
\usepackage{xcolor}
\newcommand{\promptfilebox}[2]{%
  \begin{tcolorbox}[
    enhanced,
    breakable,
    colback=gray!8,
    colframe=gray!55,
    colbacktitle=gray!30,
    coltitle=black,
    fonttitle=\bfseries\Large,
    title={#1},
    boxrule=0.5pt,
    arc=2mm,
    left=8pt,
    right=8pt,
    top=8pt,
    bottom=8pt,
    before skip=10pt,
    after skip=10pt
  ]
  \VerbatimInput[
    fontsize=\footnotesize,
    breaklines=true,
    breaksymbolleft={},
    breaksymbolright={}
  ]{#2}
  \end{tcolorbox}
}

\begin{document}

\title{Code as Representation: A Compilable Parsing Paradigm for Academic Documents}


\author{Rihui Jin}
\authornote{These authors contributed equally to this work.}
\author{Jun Wang}
\authornotemark[1]
\affiliation{
  \institution{Southeast University}
  \department{School of Computer Science and Engineering}
  \city{Nanjing}
  \state{Jiangsu} 
  \country{China}}
\email{ari\_king@seu.edu.cn}
\email{213211446@seu.edu.cn}

\author{Chengyuan Zhu}
\author{Liang Mingyu}
\author{Yue Gao}
\author{Li Yunxuan}
\affiliation{
  \institution{Southeast University}
  \department{School of Computer Science and Engineering}
  \city{Nanjing}
  \state{Jiangsu}
  \country{China}}

\author{Kuicai Dong}
\affiliation{
  \institution{Nanyang Technological University}
  \city{Singapore}
  \country{Singapore}
}
\email{kuicai001@e.ntu.edu.sg}

\author{Guilin Qi}
\authornote{Corresponding author.}
\affiliation{
  \institution{Southeast University}
  \department{School of Computer Science and Engineering}
  \city{}
  \state{}
  \country{}
}
\affiliation{
  \institution{Nanjing University}
  \department{State Key Laboratory for Novel Software Technology}
  \city{Nanjing}
  \state{Jiangsu}
  \country{China}
}

\email{gqi@seu.edu.cn}

\author{Lin Ren}
\author{Yongrui Chen}
\author{Xinbang Dai}
\author{Jiaqi Li}
\affiliation{
  \institution{Southeast University}
  \department{School of Computer Science and Engineering}
  \city{Nanjing}
  \state{Jiangsu}
  \country{China}
}

\author{Tongtong Wu}
\author{Gholamreza Haffari}
\affiliation{
    \institution{Monash University}
    \city{Clayton}
    \state{Victoria}
    \country{Australia}
}
\email{tongtong.wu@monash.edu}
\email{gholamreza.haffari@monash.edu}

\renewcommand{\shortauthors}{Rihui Jin et al.}

\renewcommand{\shortauthors}{Rihui Jin et al.}

\begin{abstract}

Academic papers are a primary carrier of scientific knowledge, yet most of this knowledge remains locked in PDFs that are optimized for human reading rather than machine use. 
For Multimodal Large Language Models (MLLMs), the core challenge is not only perception, but representation: scientific pages interleave text with Structured Academic Elements (SAEs) such as tables, formulas, charts, and pseudocode, whose structure, data, and logic are poorly preserved by common surrogates like Markdown. 
We therefore propose Compilable Academic Document Parsing (CADP), a paradigm that reconstructs a full page as contextual \LaTeX{} plus executable Python, so that structure-preserving elements and executable chart representations can be reconstructed, recompiled, and directly verified against the source page. 
To support this setting, we introduce \textsc{CADP-Bench}, an expert-verified benchmark of full academic pages containing tightly coupled text and multiple SAE types, evaluated through a re-injection compilation protocol. 
We further study current capabilities using SOTA MLLMs and an exploratory multi-agent baseline that incorporates common agentic techniques. Results show that even frontier models still struggle to produce high-fidelity executable reconstructions, highlighting substantial room for improvement in structure-aware scientific document parsing.
\textsc{CADP-Bench} is released for future research.
\footnote{\url{https://github.com/AriKing11/CADP-Bench}}

\end{abstract}


\maketitle

\section{Introduction}

Academic papers are a primary carrier of scientific knowledge, yet most of this knowledge remains locked within PDFs optimized for human reading rather than machine use~\cite{paddleocr, Huang2022LayoutLMv3PF, mmlayout}.
As Multimodal Large Language Models (MLLMs) are increasingly deployed in retrieval, reasoning, and AI-for-science pipelines~\cite{chen2025ai4research,amini2025distributed,qin2025survey,li2025unisvg}, converting scientific pages into machine-actionable representations has become a central bottleneck~\cite{dong2025doc,VRDU_survey}.
This challenge is especially acute for academic documents, where continuous prose is tightly interleaved with tables, formulas, charts, and pseudocode, which we collectively term Structured Academic Elements (SAEs).
Recovering such pages at full fidelity is therefore not only a perception problem, but fundamentally a representation problem.

\begin{figure*}
    \centering
    \includegraphics[width=0.95\linewidth]{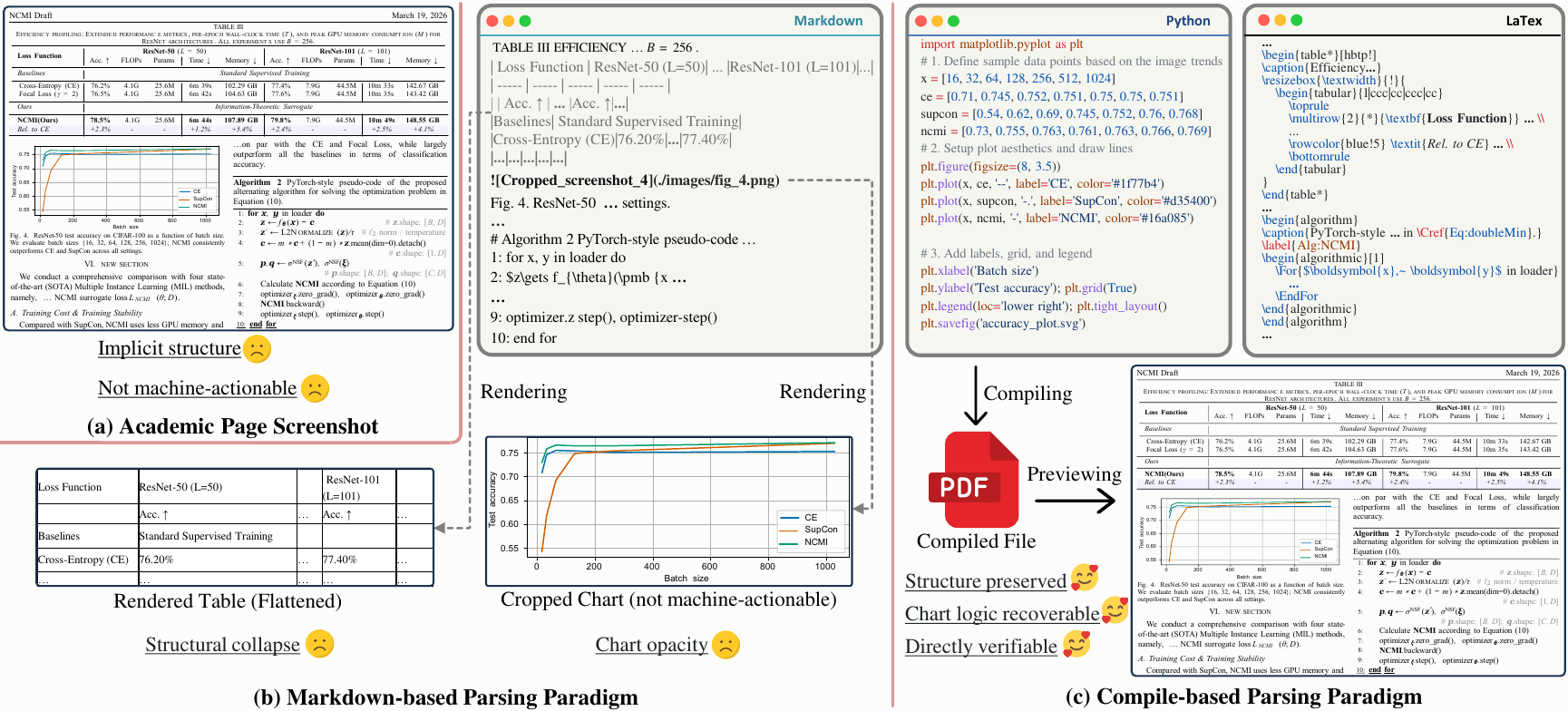}
    \vspace{-0.3cm}
    \caption{Comparison of three representations for machine reading of academic papers. (a) A raw page screenshot preserves the full visual layout but leaves structure implicit. (b) Markdown-based parsing linearizes the page into plain text, flattened tables, and raster chart crops, leading to structural collapse and opaque visual elements. (c) Our compilable parsing paradigm reconstructs the same page as contextual \LaTeX{} plus executable Python, which can be compiled back into a preview PDF, preserving document structure, recovering chart logic, and enabling direct verification.}
    \label{fig:task}
    \vspace{-0.25cm}
\end{figure*}

Downstream AI systems rarely access a paper as editable source; they usually encounter only PDFs or screenshots~\cite{rag_survey, chen2025researchpulse}.
In this process, SAE structure becomes implicit: topology, alignment, nesting, and chart data are no longer available as symbolic objects that can be directly edited, executed, or verified.
Practical pipelines therefore translate page images into machine-actionable surrogates, most commonly Markdown, because it is lightweight and LLM-friendly~\cite{idp2025, Li2025MonkeyOCRDP}.
However, Markdown is merely a convenient surface format, not a faithful representation of complex scientific pages.
It inherently suffers from three critical limitations (see Fig.~\ref{fig:task}):
\emph{structural collapse}, as its flat syntax cannot cleanly encode merged tables~\cite{jin2025hegta}, aligned equations, or complex pseudocode~\cite{tablelatex-rl};
\emph{chart opacity}, as charts are reduced to static cropped images that discard underlying data and rendering logic;
and \emph{non-verifiability}, as the parsed flat text cannot be recompiled to visually validate its fidelity against the original document.
Thus, this motivates a different question: \emph{can we recover from page pixels a representation that preserves structure, exposes chart data and rendering logic, and can be compiled back into a faithful page rendering, thereby serving as a machine-actionable substitute for the screenshot itself?}

We argue that it can, and advocate a compilable parsing paradigm that reframes parsing from pixels to programs.
We instantiate this idea as the \textbf{Compilable Academic Document Parsing} (CADP) task: given a raw full-page screenshot and limited compilation context, the model jointly generates contextual \LaTeX{} for continuous text and SAEs, and executable Python for the data and rendering logic behind charts.
This dual-code formulation preserves nested structure, makes chart internals recoverable, and, crucially, is directly verifiable: generated code can be re-injected into the original document environment, compiled, rendered, and visually compared against the source page.

Despite growing interest in document parsing (DP), existing benchmarks cannot evaluate this paradigm.
Markdown-based benchmarks~\cite{READOC,zhao2024omnidocbench} inherit the ceiling of flat representations, while element-level code generation~\cite{cat2025visual} evaluates isolated regions without page context, creating ``semantic orphans'' that lack cross-references and narrative grounding (see Table~\ref{tab:benchmark_comparison})~\cite{zhou2026scan,scimdr2026}.
To bridge this gap, we introduce \textsc{CADP-Bench}, an expert-verified benchmark for compilable academic parsing.
Each sample is a full academic page in which main text is tightly coupled with at least two SAE types, and evaluation follows a re-injection compilation protocol that compares the rendered page with the ground truth.

Using \textsc{CADP-Bench}, we evaluate SOTA MLLMs and an exploratory multi-agent baseline designed to probe their performance upper bounds.
Results show that even with agentic scaffolding, current models still struggle with high-fidelity executable reconstruction, leaving substantial room for improvement.

In summary, the main contributions of this paper are as follows:
\begin{itemize}[leftmargin=1.5em, nosep]
\item We formalize CADP, a new paradigm reconstructing pages as contextual \LaTeX{} and Python, shifting parsing to structure-preserving, dual-code generation under compilation constraints.
\item We introduce \textsc{CADP-Bench}, an expert-verified benchmark operationalizing this paradigm with multi-SAE pages, rigorously evaluated via a novel re-injection compilation protocol.
\item We benchmark SOTA MLLMs alongside an exploratory agentic baseline, and further demonstrate the dataset's broader utility for evaluating format-sensitive comprehension. Results expose key bottlenecks, confirming that high-fidelity reconstruction remains a formidable challenge.
\end{itemize}
\begin{table*}[t]
	
\centering
\newcommand{\notcheckmark}{{$\surd$}\textsuperscript{\textcolor{red}{\kern-0.35em{\bf--}}}}

\footnotesize
\setlength{\tabcolsep}{3.8pt}
\renewcommand{\arraystretch}{1.15}
\begin{tabular}{lccccccccccc}
    \toprule
    \multirow{2}{*}{Benchmarks}
                                    & \multicolumn{5}{c}{Input Type}
                                    & \multicolumn{3}{c}{Output Type}
                                    & \multirow{2}{*}{\makecell{Re-renderable                                                                                  \\Reconstruction}}
                                    & \multirow{2}{*}{\#Samples}
                                    & \multirow{2}{*}{\makecell{Annotation                                                                               \\Type}} \\
    \cmidrule(lr){2-6}
    \cmidrule(lr){7-9}
                                    & Page                                 & Table  & Chart & Formula & Pseudocode
                                    & Markdown                             & Python & \LaTeX{}
                                    &                                      &        &                                                                    \\
    \midrule
    TabLeX\cite{tablex}             & \xmark                               & \cmark & \xmark & \xmark & \xmark & \xmark & \xmark & \cmark & \cmark & 4M+  & Auto-extracted \\
    Tab-To-Tex\cite{tab-to-tex}     & \xmark                               & \cmark & \xmark & \xmark & \xmark & \xmark & \xmark & \cmark & \cmark & 40k+ & Auto-extracted \\
    TAB2LATEX\cite{latte}           & \xmark                               & \cmark & \xmark & \cmark & \xmark & \xmark & \xmark & \cmark & \cmark & 5,000 & Auto-extracted \\
    Table2LaTeX\cite{tablelatex-rl} & \xmark                               & \cmark & \xmark & \xmark & \xmark & \xmark & \xmark & \cmark & \cmark & 1,211 & Auto-extracted \\
    Im2LaTeX-100K~\cite{Img2LaTeX}      & \xmark                               & \xmark & \xmark & \cmark & \xmark & \xmark & \xmark & \cmark & \cmark & 100K & Auto-extracted \\
    UniMER~\cite{wang2024unimernetuniversalnetworkrealworld}             & \xmark                               & \xmark & \xmark & \cmark & \xmark & \xmark & \xmark & \cmark & \cmark & 20K+ & Human \\
    ChartMimic\cite{chartmimic}     & \xmark                               & \xmark & \cmark & \xmark & \xmark & \xmark & \cmark & \xmark & \cmark & 4,800 & Human          \\
    ChartX\cite{chartx}             & \xmark                               & \xmark & \cmark & \xmark & \xmark & \xmark & \cmark & \xmark & \cmark & 6,000 & LLM + Human    \\
    ChartEdit\cite{chartedit}       & \xmark                               & \xmark & \cmark & \xmark & \xmark & \xmark & \cmark & \xmark & \cmark & 1,405 & LLM + Human    \\
    DaTikZ$_{v3}$\cite{tikzero}     & \xmark                               & \xmark & \cmark & \xmark & \xmark & \xmark & \xmark & \cmark & \cmark & 1,000 & Auto + Human   \\
    OmniDocBench\cite{omnidocbench} & \cmark                               & \cmark & \cmark & \cmark & \xmark & \cmark & \xmark & \notcheckmark & \xmark & 1,355 & LLM + Human    \\
    olmOCR-bench\cite{olmocr}       & \cmark                               & \cmark & \cmark & \cmark & \xmark & \cmark & \xmark & \xmark & \xmark & 1,402 & LLM            \\
    READOC\cite{READOC}             & \cmark                               & \cmark & \cmark & \cmark & \xmark & \cmark & \xmark & \xmark & \xmark & 3,576 & Auto-extracted           \\
    \hline
    CADP-Bench (Ours)                & \cmark                               & \cmark & \cmark & \cmark & \cmark & \xmark & \cmark & \cmark & \cmark & 1,630 & LLM + Human    \\
    \bottomrule
\end{tabular}
\\
\caption{Comparison of representative document parsing benchmarks across key dimensions. \cmark: fully matches this category; \notcheckmark: partially matches this category; \xmark: does not match this category.}
\vspace{-0.35cm}
\label{tab:benchmark_comparison}

\end{table*}

\vspace{-0.2cm}
\section{Related Work}

\subsection{Multimodal Document Parsing}
Recent advancements in document digitization have been driven by both robust industrial multi-stage pipelines and end-to-end generative models~\cite{mineru, dp_unveiled, olmocr, docowl,li2025dream}. 
However, existing document parsing pipelines remain largely text-centric. Even when they incorporate localized HTML or \LaTeX{} to represent complex elements, their primary output relies on flat Markdown~\cite{idp2025, READOC}, treating non-textual graphical regions as inert raster crops. 
Consequently, much of the structural and semantic information encoded in these visuals is discarded, making the page reconstruction process inherently lossy and limiting the fidelity of the output.

Recognizing the limitations of character-level transcripts, a recent trend has shifted towards recovering document elements as structured, executable code.
Specialized models now translate isolated mathematical formulas into strict and compilable \LaTeX{}~\cite{latte, Img2LaTeX}, reverse-engineer hierarchical tables into verifiable code~\cite{tablelatex-rl, tab-to-tex, tablex}, or transpose data charts into programmatic Python scripts~\cite{tikzero, chartedit, chartllama, onechart}. 
Despite this progress, these code-generation approaches are severely constrained by an isolated cropping paradigm. 
When presented with a full screenshot of a complex academic paper, they fail to generate the comprehensive, compilable code necessary to faithfully reconstruct the original page.

\vspace{-0.2cm}

\subsection{Benchmarks for Multimodal Document Parsing}
Current DP benchmarks predominantly follow two paradigms(see Table~\ref{tab:benchmark_comparison}, each constrained by systematic limitations. 
The first paradigm focuses on flat text extraction, typically adopting Markdown as the main target representation while relegating charts and complex tables to mere links referencing cropped screenshots~\cite{omnidocbench, zhao2024omnidocbench, READOC}. 
By treating information-dense scientific charts as opaque pixels, these benchmarks fail to evaluate a model's capacity to recover underlying data arrays or logical rendering processes, resulting in a profound loss of computational utility. 
The second paradigm targets structured code generation for specific elements~\cite{chartmimic, chartx, cat2025visual}, yet it relies exclusively on localized, pre-cropped image inputs. 
Evaluating these elements in strict isolation creates ``semantic orphans.''
This localized approach fails to assess how effectively a model grounds a chart or algorithm within the broader full-page context, nor does it measure the model's ability to resolve crucial in-text cross-references.
\section{The \textsc{CADP-Bench} Benchmark}

Below, we formalize the task (\S\ref{sec:formulation}), detail dataset construction (\S\ref{sec:data}, \S\ref{sec:chart}), describe the evaluation protocol and metrics (\S\ref{sec:eval}, \S\ref{sec:metrics}), and present benchmark statistics and quality assurance (\S\ref{sec:stats}).

\vspace{-0.1cm}
\subsection{Task Formulation}
\label{sec:formulation}

\subsubsection{Input Space and Contextual Environment}
Let $\mathcal{D}$ denote a complete academic document. For a target page $p$ in $\mathcal{D}$, let
$I_p \in \mathbb{R}^{H \times W \times 3}$
denote its page image.

For re-injection compilation and evaluation, we retain a \textit{Source Context}
$\mathcal{C}_{\setminus p}$, i.e., the original \LaTeX{} source with the content of page $p$ excised.
It preserves the document-level environment, including the preamble, custom macros, references, bibliography, and surrounding text, and is used only to re-inject the generated content and verify compilation consistency.

Importantly, $\mathcal{C}_{\setminus p}$ is not provided to the model during generation.
The model receives only the target page image $I_p$ and a limited compilation context
$\mathcal{C}_{\mathrm{pkg}}$ containing the available package declarations from the document template, while additional packages may be specified when necessary.
Thus, CADP is a page-level reconstruction task under limited compilation context, rather than full source-code recovery.

\vspace{-0.1cm}
\subsubsection{Output Space: Dual-Code Generation}
The objective is to learn
$f_\theta:(I_p,\mathcal{C}_{\mathrm{pkg}})\rightarrow(\mathcal{L}_p,\mathcal{P}_p)$,
where the outputs jointly reconstruct the target page:

\begin{itemize}[leftmargin=1.5em, nosep]

\item \textbf{Contextual \LaTeX{} Block ($\mathcal{L}_p$):}
A \LaTeX{} body block that reconstructs the page text and structured elements, including tables, formulas, and pseudocode.

\item \textbf{Executable Python Programs ($\mathcal{P}_p$):}
A set of Python programs
$\mathcal{P}_p=\{P_1,P_2,\dots,P_k\}$
for reconstructing the $k$ charts on page $p$.
Rather than recovering the original authors' plotting programs or unique raw data, these programs encode visually inferred quantitative values and rendering logic, providing an executable representation that can be rendered and verified.

\end{itemize}

\begin{figure*}
\centering
\includegraphics[width=0.9\linewidth]{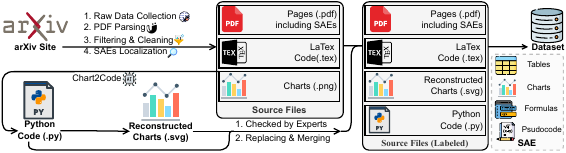}
\vspace{-0.25cm}
\caption{Construction workflow of CADP-Bench. We collect and filter arXiv source files to identify pages containing dense SAEs. Gemini-3-Pro generates candidate Python programs from raster charts, which are rendered as SVGs and retained only after expert verification of their visual content, data trends, and layout. The verified programs serve as executable chart references rather than the original plotting code. Together with the corresponding \LaTeX{} source, they form the final benchmark.}
\label{fig:benchmark_construction}
\vspace{-0.2cm}
\end{figure*}

\subsection{Data Collection and Preparation}
\label{sec:data}
As illustrated in Fig.~\ref{fig:benchmark_construction}, the construction of CADP-Bench begins with large-scale data acquisition from the arXiv repository, covering a diverse set of academic disciplines. Both PDF documents and their corresponding \LaTeX{} source files are collected to ensure comprehensive access to structural and textual information.
To filter PDFs, we employ the Mineru framework~\cite{mineru}, which extracts page-level layout information and identifies individual elements such as tables, charts, formulas, and pseudocode. Simultaneously, the collected \LaTeX{} source fragments are consolidated into complete, compilable \LaTeX{} files. Any \LaTeX{} files that fail to compile are discarded to maintain dataset integrity.

To ensure that CADP-Bench presents a sufficiently challenging benchmark aligned with the multi-SAE coupling principle, we filter for pages containing multiple SAEs. 
Specifically, using the parsing results from Mineru, we retain only those pages in which at least two distinct SAE categories coexist. 
Pages that do not meet this criterion are excluded from the benchmark.
If a single table spans two consecutive pages, we still collect this page pair as one sample even when a table is the only SAE type on those pages. This setting preserves long-range structural dependencies and evaluates whether models can recover complete tabular content under cross-page continuity.

Finally, for the selected pages, we extract the ground-truth \LaTeX{} code corresponding to all included SAEs. This extraction leverages both regular expression matching and the Mineru parsing results, ensuring precise alignment between the visual elements in the PDF and their corresponding structural representations in \LaTeX{}. The resulting dataset provides a robust foundation for evaluating full-page, multi-element reconstruction models.

\subsection{Chart Reconstruction and Annotation}
\label{sec:chart}
While tables, formulas, and pseudocode are inherently represented in \LaTeX{} source files, charts present a unique challenge: the original \LaTeX{} source typically does not contain any code that reproduces the chart, only a reference to a pre-rendered image file. This gap motivates the dual-code design of our paradigm---to achieve fully compilable reconstruction, chart data and rendering logic must be recovered as executable Python programs.

For each page retained in the benchmark, any embedded chart is first cropped and saved as a PNG image. This image is then input to a Chart2Code model---specifically, Gemini-3-Pro---which produces Python code capable of reproducing the chart. The generated Python code is subsequently compiled to render a reconstructed chart in SVG format.
To ensure the accuracy and reliability of the dataset, the reconstructed chart replaces the original image only after expert verification. 
Human annotators review the rendered chart to confirm that it faithfully reflects the original visual information, mitigating potential discrepancies introduced by the automated Chart2Code generation. 
The verified Python code serves as the ground-truth annotation, and the reconstructed chart guarantees consistency between visual representation and executable logic.

Additionally, both automated and expert-guided annotation procedures are employed to assign difficulty levels---simple, medium, or hard---to each page and its constituent SAEs. 
The difficulty is determined based on factors such as the length of the \LaTeX{} code and the structural complexity of the contained elements. Detailed criteria for this classification are provided in the Appendix.

\subsection{Evaluation Protocol: Re-injection Compilation}
\label{sec:eval}

The CADP enables an evaluation protocol that goes beyond surface-level text comparison: we test whether generated code can reproduce the original page through compilation.
Concretely, the evaluation enforces strict structural, visual, and executable consistency through a \textit{Re-injection Compilation} mechanism. 

First, each generated Python program $P_i \in \mathcal{P}_p$ is executed in an isolated environment to synthesize a SVG file $G_i$:
\begin{equation}
    G_i = \mathrm{Execute}(P_i), \quad \forall i \in \{1, \dots, k\}
\end{equation}
The generated \LaTeX{} block $\mathcal{L}_p$ must explicitly reference these generated assets (e.g., via \verb|\includegraphics{...}|).

Subsequently, $\mathcal{L}_p$ is structurally re-injected into the excision point of the source context $\mathcal{C}_{\setminus p}$, which provides the original preamble and surrounding document text. The complete, restored document is then processed by the \LaTeX{} compiler $\Omega$:
\begin{equation}
    \hat{I}_p = \Omega \big(\mathcal{C}_{\setminus p} \oplus \mathcal{L}_p, \ \{G_1, \dots, G_k\} \big)
\end{equation}
where $\oplus$ denotes in-place sequence concatenation, and $\hat{I}_p$ is the newly rendered target page. 
This process ensures that compiling the document reproduces the page content corresponding exactly to the input screenshot. 
Any hallucination of custom macros, failure to close nested environments, or incorrect data logic in Python will directly precipitate a compilation failure or severe visual misalignment, providing an inherently stringent fidelity check that is unique to the compilable parsing paradigm. 

\subsection{Evaluation Metrics}
\label{sec:metrics}
Because code is non-unique representation, we evaluate both \textit{code-level} correctness and \textit{visual-level} fidelity. 
A full summary is given in Table~\ref{tab:metric_summary}. Metrics not redefined here follow Mineru~\cite{mineru}.
Any sample that fails re-injection compilation is assigned a score of 0.

\paragraph{Pseudocode Reconstruction Score.}
We define \textit{Pseudocode Reconstruction Score} (PRS) to jointly measure textual and control-flow fidelity.
Ground-truth and predicted pseudocode are converted into ordered step sequences $S^{\mathrm{gt}}$ and $S^{\mathrm{pred}}$ (lengths $n$ and $m$), with step tokens $s^{\mathrm{gt}}_i$ and $s^{\mathrm{pred}}_i$. Preprocessing removes pseudocode tags (e.g., \verb|\State|), normalizes case/whitespace, and strips trailing punctuation. Steps are aligned strictly by order: step $i$ is compared only to step $i$; if $i>m$, set $s^{\mathrm{pred}}_i=\emptyset$; predicted steps beyond $n$ are ignored. Text fidelity is computed by normalized Levenshtein similarity:
\begin{equation}
R_{\text{text}} = \frac{1}{n}\sum_{i=1}^{n}\left(1 - \frac{\operatorname{EditDist}(s^{\mathrm{gt}}_i, s^{\mathrm{pred}}_i)}{\max(|s^{\mathrm{gt}}_i|, |s^{\mathrm{pred}}_i|)}\right).
\end{equation}
For structure fidelity independent of wording, we parse each pseudocode block into a simplified AST: the root is \texttt{ALGORITHM}; control-flow commands (\verb|\For|, \verb|\While|, \verb|\If|, \verb|\Else|, \verb|\Return|) are mapped to normalized structural nodes; and all other statements collapse into a generic \texttt{STEP}, ignoring variable names and conditions. Let $T^{\mathrm{gt}}$ and $T^{\mathrm{pred}}$ be the resulting trees. Structure fidelity is measured by normalized tree-edit similarity:
\begin{equation}
R_{\text{struct}} = \max\left(0,\ 1 - \frac{\operatorname{TED}(T^{\mathrm{gt}}, T^{\mathrm{pred}})}{\max(|T^{\mathrm{gt}}|, |T^{\mathrm{pred}}|)}\right).
\end{equation}
The final PRS averages the two components and rescales to $[0,100]$:
\begin{equation}
\operatorname{PRS} = 50\left(R_{\text{text}} + R_{\text{struct}}\right).
\end{equation}

\paragraph{Visual Reconstruction Fidelity (VRF)}
For holistic visual quality, we introduce an MLLM-as-a-judge~\cite{judge_survey} metric termed \textit{Visual Reconstruction Fidelity} (VRF) to evaluate layouts, tables, charts, and pseudocode (formulas are evaluated by separate code-level metrics). Each target type is evaluated via a tailored prompt with $K$ dimensions (e.g., completeness, structure, alignment, and visual fidelity), and each dimension is scored on a five-level ordinal scale $s_k \in \{0,1,2,3,4\}$. 
Gemini-3-Pro serves as the evaluator, having achieved an 86\% exact agreement rate against human ratings on 100 validation samples. To further assess evaluator dependence, we re-evaluate Chart VRF-A using GPT-5.5. The resulting scores differ from the Gemini-3-Pro-based evaluation by less than 3\%, while preserving the overall performance trend and Gemini-3-Pro as the best-performing model.

To enforce rigorous evaluation and prevent partial scores from overly inflating results on inherently structured elements, we employ a strict variant, \textit{VRF-S}, for tables and pseudocode. A sample scores 100 if and only if all dimensions receive a perfect mark; otherwise, it scores 0:
\begin{equation}
\operatorname{VRF-S}(x) = \begin{cases} 
100, & \text{if } s_k = 4 \text{ for all } k \in \{1, \dots, K\} \\
0, & \text{otherwise}
\end{cases}
\end{equation}
Conversely, for charts and global layout---where deterministic correctness is less binary---we define a continuous variant, \textit{VRF-A}, by averaging the individual dimensions:
\begin{equation}
\operatorname{VRF-A}(x) = 25 \cdot \frac{1}{K}\sum_{k=1}^{K} s_k
\end{equation}
The final benchmark score for a given category is obtained by averaging either $\operatorname{VRF-S}(x)$ or $\operatorname{VRF-A}(x)$ over all samples in the dataset, effectively guaranteeing precise evaluation while retaining flexibility where continuous matching is necessary.

\paragraph{Pixel Similarity.}
Let $G$ be the ground-truth target page and $B$ the blank template. If the predicted PDF renders to $T$ pages, denoted by $\hat{P}^{(t)}$ for $t=1,\dots,T$, we extend the single-page ground truth to
\begin{equation}
\tilde{G}^{(1)}=G,\qquad \tilde{G}^{(t)}=B\ \text{for } t=2,\dots,T.
\end{equation}
For each page $t$, we ignore co-background pixels and define
\begin{equation}
\mathcal{V}^{(t)}=\{(i,j):\tilde{G}^{(t)}_{ij}\neq B_{ij}\ \lor\ \hat{P}^{(t)}_{ij}\neq B_{ij}\}.
\end{equation}
Then
\begin{equation}
\Delta^{(t)}=\frac{1}{3|\mathcal{V}^{(t)}|}
\sum_{(i,j)\in \mathcal{V}^{(t)}}\sum_{c\in\{R,G,B\}}
\left|\tilde{G}^{(t)}_{ijc}-\hat{P}^{(t)}_{ijc}\right|,
\end{equation}
\begin{equation}
\mathrm{MAD}^{(t)}=\begin{cases}
0,&|\mathcal{V}^{(t)}|=0,\\
\Delta^{(t)},&|\mathcal{V}^{(t)}|>0,
\end{cases}
\qquad
\mathrm{PS}^{(t)}=1-\frac{\mathrm{MAD}^{(t)}}{255}.
\end{equation}
The sample-level score is
\begin{equation}
\mathrm{PS}=\frac{1}{T}\sum_{t=1}^{T}\mathrm{PS}^{(t)}.
\end{equation}

\begin{table}[t]
    \centering
\footnotesize
\setlength{\tabcolsep}{4pt}
\renewcommand{\arraystretch}{1.08}
\renewcommand{\tabularxcolumn}[1]{m{#1}}

\begin{tabularx}{\linewidth}{@{} >{\centering\arraybackslash}m{0.13\linewidth} >{\centering\arraybackslash}m{0.16\linewidth} >{\raggedright\arraybackslash}X @{}}
\toprule
\multicolumn{1}{c}{Type} & Metric & Notes \\
\midrule
\multirow[c]{3}{*}{Code}& Exec. Rate & Layout/Chart: re-injected \LaTeX{} and Python must  be able to be compiled and rendered. \\
\cline{3-3}
& TEDS & Table: structure\&content similarity. \\
\cline{3-3}
& PRS & Pseudocode: combines text and structure fidelity. \\
\midrule
& Reading Order & Layout: reading-order error (lower is better). \\
\cline{3-3}
\multirow[c]{5}{*}{Visual} & PageIoU & Layout: page-level overlap between prediction and ground truth. \\
\cline{3-3}
 & CDM & Formula: character-level matching score. \\
\cline{3-3}
& VRF & Layout/Table/Charts/Pseudocode: an MLLM-as-a-judge based score. \\
\cline{3-3}
& Pixel Sim. & Global pixel similarity between rendered and ground-truth pages. \\
\bottomrule
\end{tabularx}

\caption{Summary of the evaluation metrics.}
\vspace{-0.25cm}
\label{tab:metric_summary}

\end{table}

\subsection{Benchmark Statistics and Quality Assurance}
\label{sec:stats}
We summarize CADP-Bench from four complementary perspectives: overall scale, sub-domain coverage, difficulty-aware element distribution, and annotation quality.

\paragraph{Scale and Coverage.}
As shown in Table~\ref{tab:benchmark_statistics}, we report the total number of samples, the sample counts across five sub-domains, and the counts of four core SAEs under three difficulty levels. Every sample contains at least two co-occurring SAE types, ensuring consistent layout complexity across the benchmark.

\paragraph{Annotation Quality Control.}
To ensure high-fidelity ground truth, annotation and quality control were conducted by four CS undergraduate co-authors proficient in LaTeX and Python. 
Each sample was independently checked by two annotators, with disagreements adjudicated by a CS PhD candidate.
For each element type, we report the reviewer agreement rate to reflect inter-annotator consistency; the overall agreement rate is 88\%. 
Samples that do not satisfy strict alignment between the visual PDF content and the corresponding \LaTeX{}/Python ground truth are revised and re-checked before inclusion.

\begin{table}[t]
	\centering
  \footnotesize
  \renewcommand{\arraystretch}{1.12}
  \resizebox{\columnwidth}{!}{%
  \begin{tabular}{lcccccc}
    \toprule
    \textbf{Sub-domain} & 
    \begin{tabular}{@{}c@{}}Computer\\Science\end{tabular} & 
    Physics & 
    Economics & 
    \begin{tabular}{@{}c@{}}Quantitative\\Biology\end{tabular} & 
    Statistics & 
    \textbf{Total} \\
    \midrule
    \textbf{Count} & 1047 & 251 & 45 & 222 & 65 & \textbf{1630} \\
    \midrule
    \midrule
    \textbf{Element} & \textbf{Simple} & \multicolumn{2}{c}{\textbf{Medium}} & \multicolumn{2}{c}{\textbf{Hard}} & \textbf{Count} \\
    \midrule
    Table       & 803 & \multicolumn{2}{c}{644} & \multicolumn{2}{c}{395} & 1842 \\
    Formula     & 340  & \multicolumn{2}{c}{536} & \multicolumn{2}{c}{147} & 1023 \\
    Pseudocode & 40   & \multicolumn{2}{c}{71}  & \multicolumn{2}{c}{27}  & 138 \\
    Chart       & 316  & \multicolumn{2}{c}{975} & \multicolumn{2}{c}{200} & 1491 \\
    \bottomrule
  \end{tabular}%
  }
  \vspace{0.5em}
  \caption{Statistical overview of CADP-Bench, including total scale, per-sub-domain sample counts, and element-level difficulty distribution. Difficulty levels are annotated as \emph{Simple}, \emph{Medium}, and \emph{Hard}.}
  \vspace{-0.2cm}
  \label{tab:benchmark_statistics}
\end{table}

\section{Experiments \& Analysis}

\subsection{Experimental Setup}

\subsubsection{Evaluated Models}

We evaluate a set of SOTA MLLMs that represent leading approaches to full-page scientific document understanding and parsing.

\subsubsection{Exploratory Multi-Agent Baseline}
\label{sec:baseline}

\begin{figure}[t]
    \centering
    \includegraphics[width=0.93\linewidth]{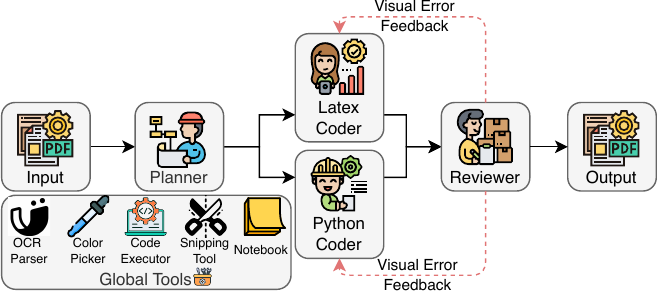}
    \vspace{-0.15cm}
    \caption{An exploratory multi-agent system used to study common agentic techniques on CADP-Bench.}
    \vspace{-0.25cm}
    \label{fig:baseline}
\end{figure}

To study whether commonly used agentic techniques are helpful on \textsc{CADP-Bench}, we implement an exploratory multi-agent system.
As shown in Fig.~\ref{fig:baseline}, it combines four common ingredients in agentic workflows: task decomposition, role specialization, shared tools, and iterative feedback. 
The system contains four agents: a Planner, a \LaTeX{} Coder, a Python Coder, and a Reviewer. Given a page image, the Planner first parses the global layout and decomposes reconstruction into text, formula, table, and chart sub-tasks; the \LaTeX{} Coder writes the page layout, text, and math; the Python Coder reconstructs chart data and plotting logic; and the Reviewer compares the compiled output with the source page and returns targeted visual feedback for another round of revision. All agents share a small tool set commonly used in document-centered agents: an OCR Parser for text and boxes, a Color Picker for chart colors, a Code Executor for \LaTeX{}/Python validation, a Snipping Tool for local crops, and a shared Notebook for intermediate notes and parameters.

\subsubsection{Implementation Details}

To reduce variability, each sample is evaluated three times, and the reported metrics are averaged across runs. All models are accessed through their officially provided APIs. The model temperature is set to 0.7. All \LaTeX{} compilations are performed using the pdfLaTex engine to ensure consistent rendering of complex structures and non-standard fonts.

\subsection{Overview}
\begin{itemize}[leftmargin=1.5em, nosep]
\item \textbf{RQ1 (End-to-End Performance):} How effectively do current SOTA MLLMs perform on \textsc{CADP-Bench}?
\item \textbf{RQ2 (Complexity \& Robustness):} How does structural difficulty, at both element and page levels, affect model reconstruction fidelity?
\item \textbf{RQ3 (Impact of Agent Techniques):} How do agent-related strategies—including self-reflection, environmental feedback, and multi-agent collaboration—affect model performance?
\item \textbf{RQ4 (Format-Sensitive Comprehension):} How does input format (screenshots, Markdown, or \LaTeX{}) influence a model's ability to answer context-dependent questions requiring information from one or more SAEs?
\end{itemize}

\subsection{Main Results (RQ1 \& RQ2)}

\begin{table*}[t]
    \centering
\setlength{\tabcolsep}{0pt}
\renewcommand{\arraystretch}{1.15}
\definecolor{visbg}{RGB}{250, 253, 248}
\definecolor{codebg}{RGB}{248, 251, 255}
\newcolumntype{C}{>{\columncolor{codebg}}c}
\newcolumntype{V}{>{\columncolor{visbg}}c}

\resizebox{\textwidth}{!}{%
\begin{tabular*}{\textwidth}{@{\extracolsep{\fill}} l V C V C V C V C V V V V @{}}
	\toprule
	\multirow{3}{*}[-1em]{Models}
	& \multicolumn{1}{c}{Formula}
	& \multicolumn{2}{c}{Table}
	& \multicolumn{2}{c}{Chart}
	& \multicolumn{2}{c}{Pseudocode}
	& \multicolumn{4}{c}{Layout}
	& \multicolumn{1}{c}{Overall} \\
	\cmidrule(lr){2-2}
	\cmidrule(lr){3-4}
	\cmidrule(lr){5-6}
	\cmidrule(lr){7-8}
	\cmidrule(lr){9-12}
	\cmidrule(lr){13-13}
	& \multicolumn{1}{c}{Visual}
	& \multicolumn{1}{c}{Code}
	& \multicolumn{1}{c}{Visual}
	& \multicolumn{1}{c}{Code}
	& \multicolumn{1}{c}{Visual}
	& \multicolumn{1}{c}{Code}
	& \multicolumn{1}{c}{Visual}
	& \multicolumn{1}{c}{Code}
	& \multicolumn{3}{c}{Visual}
	& \multicolumn{1}{c}{Visual} \\
	\cmidrule(lr){2-2}
	\cmidrule(lr){3-3}
	\cmidrule(lr){4-4}
	\cmidrule(lr){5-5}
	\cmidrule(lr){6-6}
	\cmidrule(lr){7-7}
	\cmidrule(lr){8-8}
	\cmidrule(lr){9-9}
	\cmidrule(lr){10-12}
	\cmidrule(lr){13-13}
	& CDM$\uparrow$ & TEDS$\uparrow$ & VRF-S$\uparrow$ & \makecell{Exec.\\Rate}$\uparrow$ & VRF-A$\uparrow$ & PRS$\uparrow$ & VRF-S$\uparrow$ & \makecell{Exec.\\Rate}$\uparrow$ & \makecell{Reading\\Order}$\downarrow$ & PageIoU$\uparrow$ & VRF-A$\uparrow$ & \makecell{Pixel\\Sim.}$\uparrow$ \\
	\midrule

    Claude Haiku 4.5  & 27.71 & 44.10 & 9.46 & 49.57 & 4.75  & 10.82 & 0.00 & 42.33  			& 75.50 			& 23.17 			& 25.75 		& 20.32 \\
	Claude Sonnet 4.6 & 52.10 & 72.80 & 38.28 & 93.59 & 27.75 & 47.74 & 21.74 & 86.33  			& 46.03 			& 46.34 			& 50.50 		& 41.48 \\
	Claude Opus 4.6   & 60.61 & 71.32 & 35.43 & 78.63 & 35.00 & 49.50 & 17.39 & 84.00  			& 49.97 			& 49.45 			& 48.50 		& 39.76 \\
	Qwen3.5-35B-A3B   & 63.69 & 72.35 & 16.50 & 72.65 & 11.50 & 59.77 & 8.70 & 69.70  			& 57.52 			& 35.60 			& 42.75 		& 33.01 \\
	Qwen3.5-27B       & 68.38 & 84.03 & 48.25 & 71.64 & 18.75 & 65.88 & 26.09 & 80.13  			& 46.33 			& 44.76 			& 52.00 		& 39.40 \\
	Qwen3.5-122B-A10B & 69.38 & 88.31 & 54.65 & 85.39 & 43.75 & 73.10 & 13.04 & 85.19  			& 40.15 			& 50.42 			& 53.50 		& 40.57 \\
	Qwen3.5-Flash     & 64.88 & 75.48 & 38.07 & 73.35 & 13.25 & 67.36 & 4.35 & 69.36  			& 57.13 			& 36.90 			& 41.75 		& 33.01 \\
	Qwen3.5-Plus      & 69.96 & 91.92 & 42.20 & 93.01 & 40.50 & 73.67 & 21.74 & 89.23  			& 34.89 			& 55.08 			& 61.25 		& 44.18 \\
	Qwen3.5-397B-A17B & \underline{70.45} & 92.17 & 61.39 & 92.23 & 45.50 & 73.78 & 34.78 & 93.27  			& 33.86 			& 55.80 			& 64.25 		& 45.74 \\
	GPT-5.3-Codex     & 68.55 & 90.68 & 36.50 & 96.15 & 54.25 & 39.83 & 8.70 & 91.33  			& 34.82 			& 55.94 			& 58.25 		& 44.29 \\
	GPT-5.4           & 66.64 & 87.39 & 43.26 & \textbf{97.86} & \underline{59.25} & 41.26 & 13.04 & 90.00  			& 36.60 			& 53.83 			& 58.50 		& 43.54 \\
	Kimi-K2.5         & 69.51 & \underline{92.68} & 57.14 & \underline{97.60} & 43.25 & \underline{78.55} & 30.43 & \underline{93.60}  			& 32.21 			& 56.39 			& 62.75 		& \underline{45.93} \\
	Gemini-3-Flash    & 70.38 & 89.63 & \underline{62.48} & 97.01 & 52.75 & 72.88 & \underline{34.78} & 92.33  			& \underline{31.06} 			& \underline{57.78} 			& \underline{65.50} 		& 45.39 \\
	Gemini-3-Pro    & \textbf{70.83} & \textbf{94.49} & \textbf{63.09} & 96.97 & \textbf{60.00} & \textbf{80.51} & \textbf{56.52} & \textbf{94.95}  & \textbf{25.19} 	& \textbf{61.52} 	& \textbf{73.00} & \textbf{47.28} \\

	\bottomrule
\end{tabular*}

 }
\caption{Overall model performance across layout, formula, table, chart, and Pseudocode parsing tasks under code- and visual-based metrics (higher is better except Reading Order). Light green: Visual; light blue: Code.}
\vspace{-0.2cm}
\label{tab:eval_metrics1}
\end{table*}

\subsubsection{E2E Performance (RQ1)}
From Table~\ref{tab:eval_metrics1}, we summarize two key findings from the dual-code evaluation: 
1) From a capability perspective, while open-weight models are rapidly closing the gap with proprietary systems, complex structural reasoning remains a pervasive bottleneck. Gemini-3-Pro stands out as the top performer, yet its sub-optimal VRF layout and Pixel Sim. scores show that flawless high-fidelity generation remains challenging. Notably, the open-weight Qwen3.5-397B-A17B achieves overall visual fidelity rivaling commercial counterparts like GPT-5.4 and Claude Opus 4.6. However, structurally dense elements like pseudocode expose severe vulnerabilities across all capability tiers. While Gemini-3-Pro maintains a lead (56.52 VRF-S), other strong frontier models (e.g., GPT-5.4 at 13.04, Claude Opus 4.6 at 17.39) suffer dramatic performance collapses, underscoring that robust algorithmic formatting is a universally unsolved challenge for current MLLMs.
2) From a metric perspective, code executability and visual fidelity reveal a systematic decoupling in chart reconstruction. 
Frontier models achieve exceptionally high execution rates, yet their VRF-A scores lag significantly behind, exposing a fundamental gap: generating runnable plotting code is far easier for current models than faithfully reproducing the nuanced aesthetics and data distributions of the original figures. 
Similarly, conventional metrics like TEDS paint an overly optimistic picture of table parsing, while our stricter VRF-S criterion reveals a steep performance plunge. This pattern indicates that partial structural recovery is insufficient—only perfect reconstruction counts.

\subsubsection{Complexity \& Robustness (RQ2)}
\begin{figure*}
    \centering
    \vspace{-0.2cm}
    \includegraphics[width=0.94\linewidth]{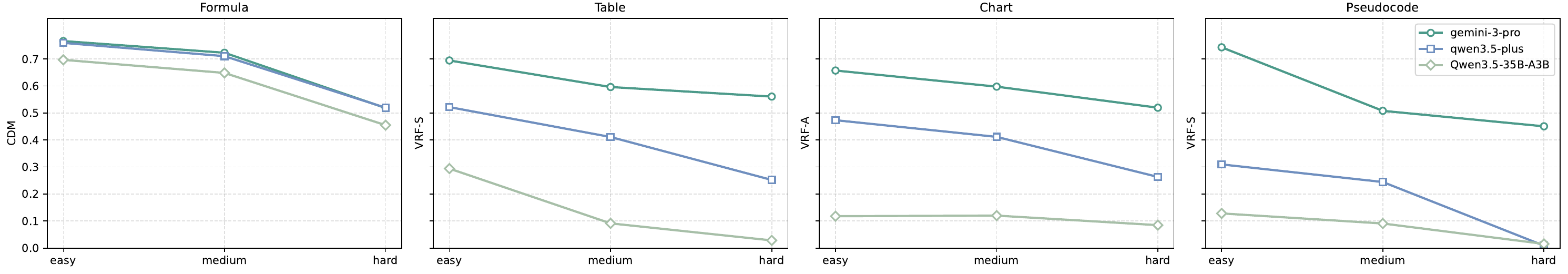}
    \vspace{-0.2cm}
    \caption{Performance across element-level difficulty (easy, medium, hard) for three representative models at different performance tiers: Gemini-3-Pro (strong), Qwen3.5-Plus (medium), and Qwen3.5-35B-A3B (weak). Table reconstruction is the most robust, while formula, chart, and pseudocode performance degrades with increasing complexity.}
    \label{fig:rq2}
    \vspace{-0.2cm}
\end{figure*}

Fig.~\ref{fig:rq2} reveals a clear difficulty-dependent degradation across all four element types, but with markedly different failure profiles by model tier. For formulas, all three models drop as difficulty increases, with Gemini-3-Pro and Qwen3.5-Plus showing relatively smooth declines, while Qwen3.5-35B-A3B falls more sharply on hard cases. 

For tables and charts, the tier gap widens substantially at higher difficulty: Gemini-3-Pro degrades moderately, whereas Qwen3.5-Plus exhibits a large hard-case drop and Qwen3.5-35B-A3B collapses to near-floor performance on tables and remains consistently low on charts. The largest instability appears in pseudocode, where both Qwen models approach near-zero on hard samples, while Gemini-3-Pro, although still strongest, also shows a pronounced decline from easy to hard. 

Overall, the curves indicate that current MLLMs retain partial robustness on formula and table reconstruction only at higher capability tiers, while chart and pseudocode reconstruction remain the dominant bottlenecks under increasing structural complexity.

\subsubsection{Case Study}
\begin{figure}
    \centering
    \includegraphics[width=0.99\linewidth]{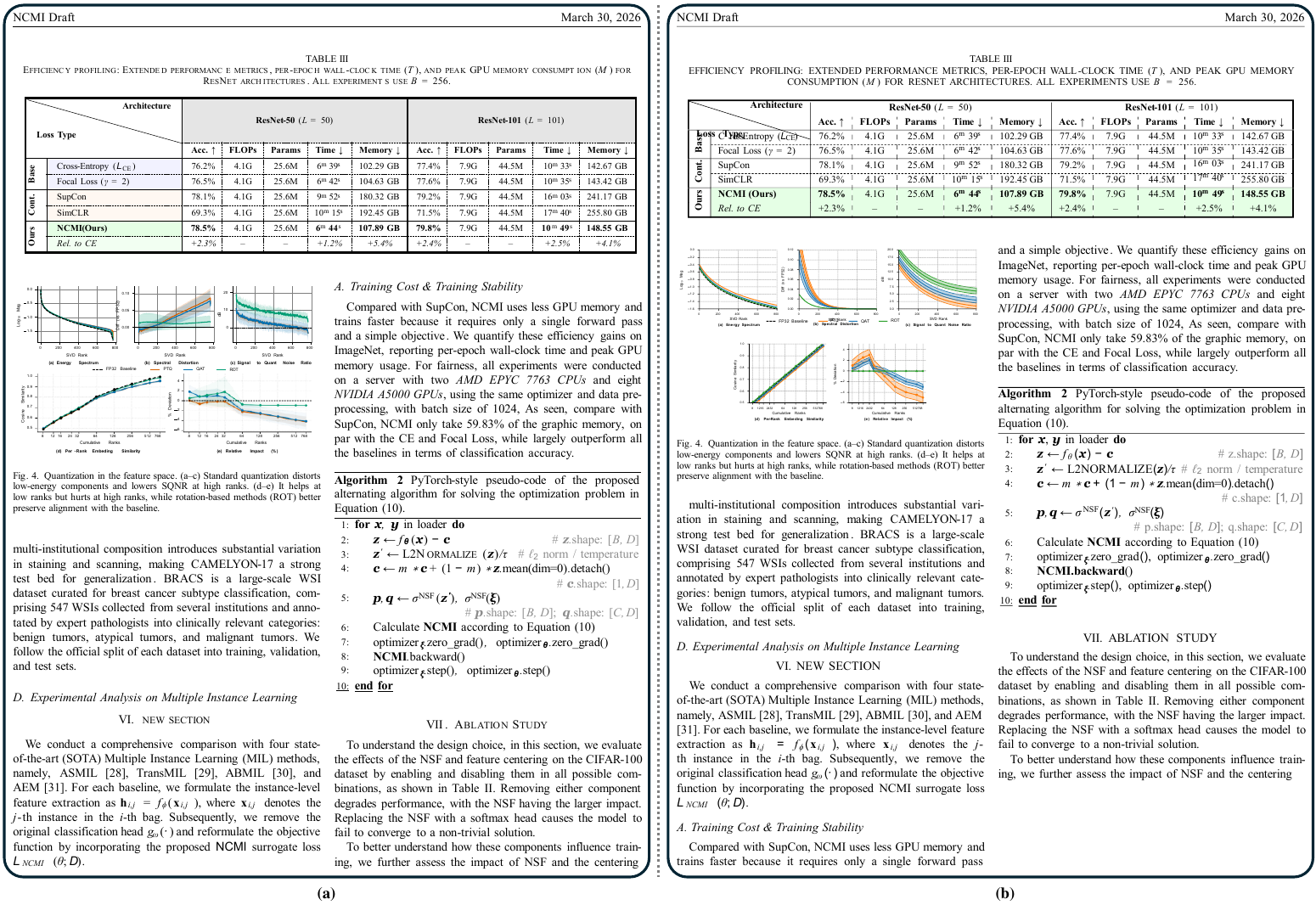}
    \vspace{-0.2cm}
    \caption{Case study of CADP on a challenging sample: the source screenshot (left panel) v.s. the parsed result by Gemini-3-Pro (right panel). VRF-A scores: Layout 75.00, Table 58.33, Chart 43.75 and Pseudocode 100. }
    \label{fig:case_study}
    \vspace{-0.2cm}
\end{figure}

Fig.~\ref{fig:case_study} shows an academic screenshot containing a challenging table and a complex chart (left), together with the parsing result produced by Gemini-3-Pro. 
As can be seen, the reconstructed table exhibits substantial deviations, particularly in its color rendering and the text in the upper-left corner. 
The chart parsing is even more problematic, with severe errors throughout. 
In addition, inconsistencies in the sizes of the reconstructed table and chart relative to the original screenshot lead to noticeable discrepancies in the overall page layout. This trend is also reflected by the VRF-A scores in this case study: Layout 75.00, Table 58.33, Chart 43.75 and Pseudocode 100.

\subsection{Impact of Agentic Techniques (RQ3)}
To answer RQ3—which explores how common agent strategies affect the model's DP capabilities—we systematically evaluate our exploratory multi-agent baseline across different configurations.
As shown in Table~\ref{tab:ablation_metrics1}, we degrade the full baseline system by incrementally removing core agentic modules (e.g., w/o Visual Feedback, w/o Tools, and w/o Multi-Agent). 
For a more comprehensive comparison, we also include a setting where the base model is augmented solely with simple self-reflection (Base w/ Self-reflection). 
By comparing these configurations, we can empirically isolate and quantify how different levels of agentic scaffolding contribute to the final full-page reconstruction accuracy, and assess whether current MLLMs can fundamentally overcome their limitations when provided with external tools and iterative reasoning.

\begin{table}[t]
    \centering
\setlength{\tabcolsep}{4pt}
\renewcommand{\arraystretch}{1.15}

\resizebox{\linewidth}{!}{%
\begin{tabular}{l c ccccc}
\toprule
\multirow{2}{*}{Setting}
& \multirow{2}{*}{\makecell{Avg.\\Calls}}
& Layout & Formula & Table & Chart & Pseudocode \\
&& (VRF-A) & (CDM) & (VRF-S) & (VRF-A) & (VRF-S) \\
\midrule
Base (Gemini-3-Pro) & \textbf{1.0} & 73.00 & 70.83 & 66.09 & 60.00 & 56.52 \\
\hspace{0.6em}w/ Self-reflection & 2.0 & 71.00 & 68.80 & 63.99 & 58.50 & 52.17 \\

\midrule
Full MA System & 6.0 & \textbf{77.20} & 74.20 & \textbf{71.40} & \textbf{66.50} & 61.80 \\
\hspace{0.6em}w/o Visual Feedback  & 4.5 & 74.10 & \textbf{75.30} & 66.80 & 62.00 & 58.50 \\
\hspace{0.6em}w/o Tools & 5.0 & 76.50 & 73.40 & 68.20 & 61.80 & 61.10 \\
\hspace{0.6em}w/o Multi-Agent & 3.0 & 75.80 & 72.60 & 67.90 & 63.50 & \textbf{62.50} \\
\bottomrule
\end{tabular}
}

\caption{Ablation study for RQ3, designed to quantify how agentic components contribute to reconstruction quality.}
\vspace{-0.4cm}

\label{tab:ablation_metrics1}
\end{table}

Table~\ref{tab:ablation_metrics1} highlights several key insights about agentic scaffolding. 
First, the Full MA System yields a solid and consistent performance gain of roughly 3 to 7 points across all dimensions compared to the static Base model. This confirms the overall efficacy of combining tooling, feedback, and decomposition. 
However, while most modules contribute positively to complex elements like charts and layouts, we observe intriguing counter-productive effects in structurally rigid domains. For instance, removing Visual Feedback actually increases the score for Formula (from 74.20 to 75.30). This suggests that pixel-based visual critics sometimes misinterpret dense, tiny math symbols, overriding correct \LaTeX{} source with hallucinated "fixes." 
Similarly, removing Multi-Agent collaboration marginally improves Pseudocode reproduction (62.50 vs.\ 61.80). This occurs because passing raw textual code through multiple redundant generation cycles between agents inadvertently degrades precise whitespace indents and specific syntax structures.
Finally, augmenting the base model with self-reflection causes a uniform performance regression, reinforcing the fact that reflection without external environmental grounding only introduces instability.

\vspace{-0.1cm}
\subsection{Format-Sensitive Comprehension (RQ4)}

To investigate how input representation affects structured scientific comprehension, we construct a QA benchmark of 90 manually annotated questions requiring evidence from one or two SAEs and their surrounding context. 
We evaluate three MLLMs at different performance tiers (Gemini-3-Pro, Qwen3.5-Plus, and Qwen3.5-35B-A3B) under three input formats: full-page screenshots, Markdown, and our structured \LaTeX{}+Python representation.
For each query, the model receives the question and the complete paper in one of these formats.

\begin{figure}
    \centering
    \includegraphics[width=0.97\linewidth]{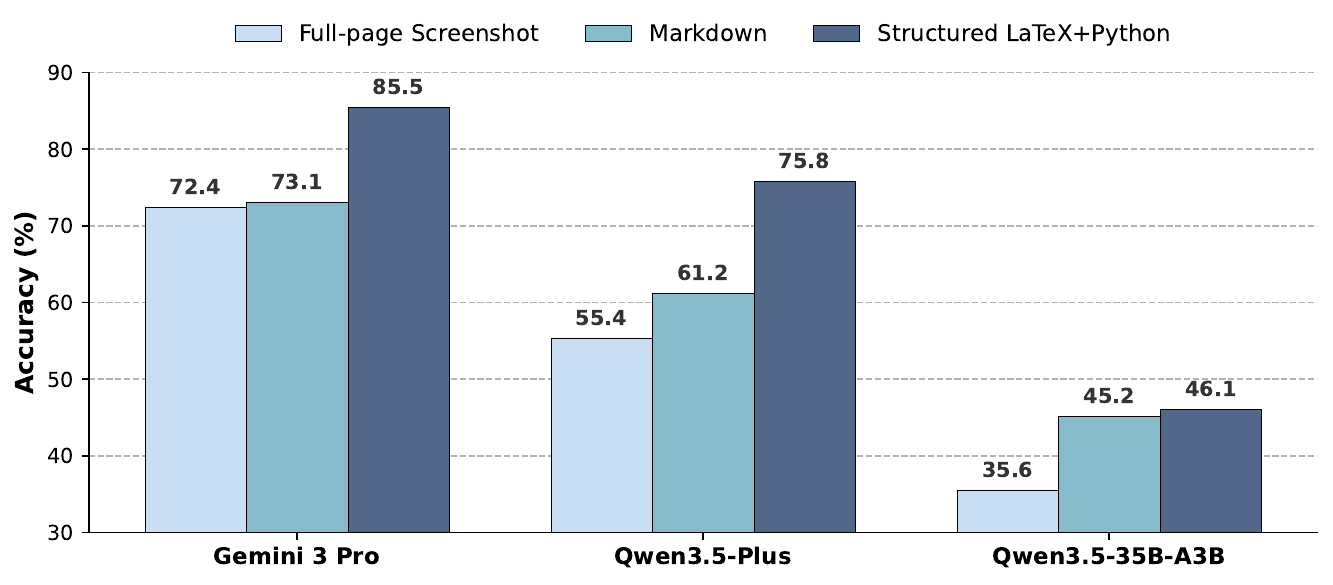}
    \vspace{-0.2cm}
    \caption{Format-sensitive QA performance across three input representations.}
    \label{fig:rq4}
    \vspace{-0.2cm}
\end{figure}

Fig.~\ref{fig:rq4} reveals a capability-dependent effect of representation. Qwen3.5-35B-A3B gains little from \LaTeX{}+Python over Markdown, suggesting that smaller models struggle to exploit complex programmatic structure. 
Gemini-3-Pro performs similarly on screenshots and Markdown, indicating that strong visual grounding can partially mitigate the lossiness of raw page images. 
In contrast, Qwen3.5-Plus benefits substantially from \LaTeX{}+Python and even surpasses Gemini-3-Pro with Markdown. 
This performance inversion suggests that structure-preserving representations provide useful inductive biases that can partially compensate for model-scale limitations and improve scientific comprehension.

\vspace{-0.15cm}

\section{Discussion}

The proposed CADP paradigm extends beyond benchmark performance and offers practical value for downstream scientific document understanding. In particular, the Dual-Code representation preserves both page-level structure and executable semantics, providing a stronger foundation than plain-text formats for full-page reconstruction and reasoning.

\textbf{Potential for Reinforcement Learning in Compilable Document Parsing.}
Compared with flat Markdown, Dual-Code is substantially more expressive for reconstructing complex academic pages. This makes CADP naturally compatible with reinforcement learning: generated code can be compiled/executed, rendered outputs can be directly compared with reference pages, and reconstruction metrics can serve as reward signals. Given the large-scale \LaTeX{} ecosystem (e.g., arXiv), this setting is promising for pretraining and post-training toward robust scientific document parsing.

\textbf{Enabling Graph-Based Retrieval-Augmented Reasoning.}
\LaTeX{} also provides explicit structural signals that are useful for Graph-RAG. Cross-references (\textbackslash ref), citations (\textbackslash cite), table hierarchies, and section organization encode rich relations across text and SAEs (e.g., tables and charts). Preserving these relations in Dual-Code can alleviate the structure-breaking issue of chunk-based RAG and improve both intra-document and cross-document QA.
\section{Conclusion}

We present CADP, a paradigm that reconstructs academic pages as contextual \LaTeX{} and executable Python, together with \textsc{CADP-Bench}, an expert-verified benchmark evaluated via re-injection compilation. Experiments on SOTA MLLMs and an exploratory multi-agent baseline show that high-fidelity executable reconstruction remains challenging. Meanwhile, structured \LaTeX{}+Python improves downstream comprehension, particularly for medium-tier models, highlighting the potential of compilable, structure-preserving representations for scientific document understanding.

\clearpage
\begin{acks}
This work is partially supported by National Nature Science Foundation of China under No. 62476058. We thank the Big Data Computing Center of Southeast University for providing the facility support on the numerical calculations in this paper.
\end{acks}

\bibliographystyle{ACM-Reference-Format}
\bibliography{sample-base}

\appendix
\onecolumn
\section{Prompts}

\promptfilebox{Prompt for Compilable Academic Document Parsing}{prompts/main_prompt.txt}

\promptfilebox{Prompt for Table Restoration Evaluation}{prompts/table_prompt.txt}

\promptfilebox{Prompt for Chart Restoration Evaluation}{prompts/chart_prompt.txt}

\promptfilebox{Prompt for Document Layout Restoration Evaluation}{prompts/layout_prompt.txt}

\promptfilebox{Prompt for Pseudocode Restoration Evaluation}{prompts/alg_prompt.txt}


\end{document}